\documentclass[twoside,leqno,twocolumn]{article}
\newif\ifapx
\apxtrue 

\usepackage{times}
\usepackage[english]{babel}

\usepackage{ltexpprt}
\usepackage{algorithmic}
\usepackage{algorithm}
\usepackage{amsmath}
\usepackage{amssymb}
\usepackage{verbatim}
\usepackage{multirow}
\usepackage{booktabs}
\usepackage{url}
\usepackage{cite}
\usepackage{xspace}
\usepackage{array}
\usepackage{dcolumn}
\usepackage{graphicx}
\usepackage[tight,center]{subfigure}
\usepackage{color}
\usepackage{makeidx}
\usepackage{verbatim}
\usepackage{caption}
\usepackage{float}
\usepackage{enumitem}
\usepackage{pifont}
\usepackage{enumitem} 
\usepackage{calc} 

\usepackage{tikz}
\usetikzlibrary{snakes,arrows,shapes,positioning,fit,calc,patterns}
\usepackage{pgfplots}

\newcommand{\otoprule }{\midrule[\heavyrulewidth]}

\newtheorem{definitionB}{Definition}
\newenvironment{definition}{\begin{definitionB}\textit\bgroup}{\egroup\end{definitionB}}

\newcommand{\proofApx}{
\begin{proof}
We postpone the proof to Appendix~\ref{sec:proofs}.
\end{proof}
}

\newcommand{\ourmethod}{\textsc{CDA}\xspace}
\newcommand{\cda}{\ourmethod}

\newcommand{\Rcda}{\textsc{rCDA}\xspace}
\newcommand{\Mcda}{\textsc{mCDA}\xspace}
\newcommand{\MRcda}{\textsc{mrCDA}\xspace}

\newcommand{\mcda}{\textsc{CDA}$_M$\xspace}
\newcommand{\mRcda}{\textsc{rCDA}$_M$\xspace}

\newcommand{\qcda}{\textsc{CDA}$_Q$\xspace}
\newcommand{\qRcda}{\textsc{rCDA}$_Q$\xspace}

\newcommand{\pecda}{\textsc{CDA}$_{\mathit{P}}$\xspace}
\newcommand{\peMRcda}{\textsc{mrCDA}$_{\mathit{P}}$\xspace}
\newcommand{\peMcda}{\textsc{mCDA}$_{\mathit{P}}$\xspace}

\newcommand{\cca}{\textsc{CCA}\xspace}
\newcommand{\lscca}{\textsc{LCCA}\xspace}

\newcommand{\kde}{\textsc{KDE}\xspace}

\newcommand{\rescu}{\textsc{RESCU}\xspace}
\newcommand{\nid}{\textsc{NID}\xspace}
\newcommand{\tca}{\textsc{TCA}\xspace}
\newcommand{\scl}{\textsc{SCL}\xspace}
\newcommand{\dac}{\textsc{DAC}\xspace}
\newcommand{\ssc}{\textsc{SSC}\xspace}

\newcommand{\D}{\mathbf{D}}
\newcommand{\E}{\mathbf{E}}
\newcommand{\U}{\mathbf{U}}
\newcommand{\V}{\mathbf{V}}
\newcommand{\I}{\mathbf{I}}
\newcommand{\e}{\mathbf{e}}
\newcommand{\Xb}{\mathbf{X}}
\newcommand{\Yb}{\mathbf{Y}}
\newcommand{\xb}{\mathbf{x}}
\newcommand{\yb}{\mathbf{y}}
\newcommand{\X}{X}
\newcommand{\Y}{Y}
\newcommand{\x}{x}
\newcommand{\y}{y}

\newcommand{\SizeX}{n}
\newcommand{\model}{\mathcal{A}}
\newcommand{\DimX}{m}
\newcommand{\SizeY}{k}
\newcommand{\DimY}{l}

\newcommand{\vecu}{u}
\newcommand{\vecv}{v}
\newcommand{\diff}{\mathit{diff}}
\newcommand{\obj}{\mathit{obj}}
\newcommand{\dist}{\mathit{dist}}

\newcommand{\dom}{\mathit{dom}}

\newcommand{\md}{\textsc{md}\xspace}
\newcommand{\qr}{\textsc{qr}\xspace}

\newcommand{\pe}{\textsc{pe}\xspace}
\newcommand{\csel}{\mathit{\mathcal{C}_{sel}}}
\newcommand{\ccand}{\mathit{\mathcal{C}_{cand}}}
\newcommand{\cost}{\mathit{cost}}
\newcommand{\cov}{\mathit{cov}}
\newcommand{\potent}{\mathit{potent}}

\makeatletter
\renewcommand*{\@fnsymbol}[1]{\ensuremath{\ifcase#1\or   \circ\or \bullet\or *\or \ddagger\or
   \mathsection\or \mathparagraph\or \|\or **\or \dagger\dagger
   \or \ddagger\ddagger \else\@ctrerr\fi}}
\makeatother

\pgfdeclarelayer{background}
\pgfdeclarelayer{foreground}
\pgfsetlayers{background,main,foreground}

\tikzstyle{block} = [rounded corners, draw=blue!70, fill=white, text width=3.3cm, minimum height=4em]
\tikzstyle{bgblock} = [rounded corners, draw=blue!70, thick, fill=blue!10, text width=3.3cm, minimum height=4em]
\tikzstyle{line} = [draw, -latex', thick,blue!70]

\definecolor{yafaxiscolor}{rgb}{0.3, 0.3, 0.3}

\definecolor{yafcolor1}{rgb}{0.4, 0.165, 0.553}
\definecolor{yafcolor2}{rgb}{0.949, 0.482, 0.216}
\definecolor{yafcolor3}{rgb}{0.47, 0.549, 0.306}
\definecolor{yafcolor4}{rgb}{0.925, 0.165, 0.224}
\definecolor{yafcolor5}{rgb}{0.141, 0.345, 0.643}
\definecolor{yafcolor6}{rgb}{0.965, 0.933, 0.267}
\definecolor{yafcolor7}{rgb}{0.627, 0.118, 0.165}
\definecolor{yafcolor8}{rgb}{0.878, 0.475, 0.686}

\newlength{\yafaxispad}
\newlength{\yaftlpad}
\newlength{\yaflabelpad}
\newlength{\yafaxiswidth}
\newlength{\yafticklen}
\makeatletter
\def\pgfplots@drawtickgridlines@INSTALLCLIP@onorientedsurf#1{}
\makeatother

\newcommand{\yafdrawaxes}[4]{
	\pgfplotstransformcoordinatex{#1}\let\xmincoord=\pgfmathresult 
	\pgfplotstransformcoordinatex{#2}\let\xmaxcoord=\pgfmathresult 
	\pgfplotstransformcoordinatey{#3}\let\ymincoord=\pgfmathresult 
	\pgfplotstransformcoordinatey{#4}\let\ymaxcoord=\pgfmathresult 
	\pgfsetlinewidth{\yafaxiswidth} 
	\pgfsetcolor{yafaxiscolor}
	\pgfpathmoveto{\pgfpointadd{\pgfpointadd{\pgfplotspointrelaxisxy{0}{0}}{\pgfqpointxy{\xmincoord}{0}}}{\pgfqpoint{-0.5\yafaxiswidth}{\yafaxispad}}}
	\pgfpathlineto{\pgfpointadd{\pgfpointadd{\pgfplotspointrelaxisxy{0}{0}}{\pgfqpointxy{\xmaxcoord}{0}}}{\pgfqpoint{0.5\yafaxiswidth}{\yafaxispad}}}
	\pgfpathmoveto{\pgfpointadd{\pgfpointadd{\pgfplotspointrelaxisxy{0}{0}}{\pgfqpointxy{0}{\ymincoord}}}{\pgfqpoint{\yafaxispad}{-0.5\yafaxiswidth}}}
	\pgfpathlineto{\pgfpointadd{\pgfpointadd{\pgfplotspointrelaxisxy{0}{0}}{\pgfqpointxy{0}{\ymaxcoord}}}{\pgfqpoint{\yafaxispad}{0.5\yafaxiswidth}}}
	\pgfusepath{stroke}
}

\newcommand{\yafdrawYaxis}[2]{
	\pgfplotstransformcoordinatey{#1}\let\ymincoord=\pgfmathresult 
	\pgfplotstransformcoordinatey{#2}\let\ymaxcoord=\pgfmathresult 
	\pgfsetlinewidth{\yafaxiswidth} 
	\pgfsetcolor{yafaxiscolor}
	\pgfpathmoveto{\pgfpointadd{\pgfpointadd{\pgfplotspointrelaxisxy{0}{0}}{\pgfqpointxy{0}{\ymincoord}}}{\pgfqpoint{\yafaxispad}{-0.5\yafaxiswidth}}}
	\pgfpathlineto{\pgfpointadd{\pgfpointadd{\pgfplotspointrelaxisxy{0}{0}}{\pgfqpointxy{0}{\ymaxcoord}}}{\pgfqpoint{\yafaxispad}{0.5\yafaxiswidth}}}
	\pgfusepath{stroke}
}

\newcommand{\yafdrawaxisLimits}[4]{
	\pgfplotstransformcoordinatex{#1}\let\xmincoord=\pgfmathresult
	\pgfplotstransformcoordinatex{#2}\let\xmaxcoord=\pgfmathresult
	\pgfplotstransformcoordinatey{#3}\let\ymincoord=\pgfmathresult 
	\pgfplotstransformcoordinatey{#4}\let\ymaxcoord=\pgfmathresult 
	\pgfsetlinewidth{\yafaxiswidth} 
	\pgfsetcolor{yafaxiscolor}
	\pgfpathmoveto{
		\pgfpointadd{
			\pgfpointadd{
				\pgfplotspointrelaxisxy{0}{0}}{
				\pgfqpointxy{\xmincoord}{0}
			}
		}{
			\pgfqpoint{
				-0.5\yafaxiswidth}{
				\yafaxispad
			}
		}
	}
	\pgfpathlineto{
		\pgfpointadd{
			\pgfpointadd{
				\pgfplotspointrelaxisxy{0}{0}
			}{
				\pgfqpointxy{
					\xmaxcoord
				}{0}
			}
		}{
			\pgfqpoint{
				25.5\yafaxiswidth
			}{
				\yafaxispad
			}
		}
	}
	\pgfpathmoveto{\pgfpointadd{\pgfpointadd{\pgfplotspointrelaxisxy{0}{0}}{\pgfqpointxy{0}{\ymincoord}}}{\pgfqpoint{\yafaxispad}{-0.5\yafaxiswidth}}}
	\pgfpathlineto{\pgfpointadd{\pgfpointadd{\pgfplotspointrelaxisxy{0}{0}}{\pgfqpointxy{0}{\ymaxcoord}}}{\pgfqpoint{\yafaxispad}{0.5\yafaxiswidth}}}
	\pgfusepath{stroke}
}

\pgfplotscreateplotcyclelist{yaf}{%
{yafcolor1,mark options={scale=0.75},mark=o}, 
{yafcolor2,mark options={scale=0.75},mark=square},
{yafcolor3,mark options={scale=0.75},mark=triangle},
{yafcolor4,mark options={scale=0.75},mark=o},
{yafcolor5,mark options={scale=0.75},mark=o},
{yafcolor6,mark options={scale=0.75},mark=o},
{yafcolor7,mark options={scale=0.75},mark=o},
{yafcolor8,mark options={scale=0.75},mark=o}} 

\pgfplotscreateplotcyclelist{yaf fill}{%
{fill=yafcolor1,draw=none}, 
{fill=yafcolor2,draw=none},
{fill=yafcolor3,draw=none},
{fill=yafcolor4,draw=none},
{fill=yafcolor5,draw=none},
{fill=yafcolor6,draw=none},
{fill=yafcolor7,draw=none},
{fill=yafcolor8,draw=none}} 

\pgfplotsset{jv ybar/.style={
   ybar, 
   cycle list name=yaf fill,
   xtick = \empty,
   every extra x tick/.style={major tick length=0pt,color=black}, 
   xmajorgrids = false,
}}

\pgfplotsset{jv line/.style={
   no markers,
   cycle list name=yaf,
   log ticks with fixed point,
   y tick label style = {/pgf/number format/set thousands separator = {\,}},
}}
\pgfplotsset{jv line ylog/.style={
   jv line,
}}

\pgfplotsset{axis y line=left, axis x line=bottom,
	tick align=outside,
	tickwidth=\yafticklen,
	clip = false,
    x axis line style= {-, line width = 0pt, color=black!0},
    y axis line style= {-, line width = 0pt, color=black!0},
    x tick style= {line width = \yafaxiswidth, color=yafaxiscolor, yshift = \yafaxispad},
    y tick style= {line width = \yafaxiswidth, color=yafaxiscolor, xshift = \yafaxispad},
    x tick label style = {font=\scriptsize, yshift = \yaftlpad},
    y tick label style = {font=\scriptsize, xshift = \yaftlpad},
    every axis y label/.style = {at = {(ticklabel cs:0.5)}, rotate=90, anchor=center, font=\scriptsize, yshift = -\yaflabelpad},
    every axis x label/.style = {at = {(ticklabel cs:0.5)}, anchor=center, font=\scriptsize, yshift = \yaflabelpad},
    x tick label style = {font=\scriptsize, yshift = 1pt},
    grid = major,
    major grid style  = {dash pattern = on 1pt off 3 pt},
	every axis plot post/.append style= {line width=\yafaxiswidth} ,
	legend cell align = left,
	legend style = {inner sep = 1pt, cells = {font=\scriptsize}},
	legend image code/.code={%
		\draw[mark repeat=2,mark phase=2,#1] 
		plot coordinates { (0cm,0cm) (0.15cm,0cm) (0.3cm,0cm) };%
	} 
}

\begin{document}

\title{Canonical Divergence Analysis}

\author{
Hoang-Vu Nguyen\thanks{Max Planck Institute for Informatics and Saarland University, Germany. Email: \texttt{\{hnguyen,jilles\}@mpi-inf.mpg.de}} \hspace{2.0cm}
Jilles Vreeken\footnotemark[1]
}

\date{}

\maketitle

\begin{abstract}
\small\baselineskip=9pt%
We aim to analyze the relation between two random vectors that may potentially have both different number of attributes as well as realizations, and which may even not have a joint distribution. This problem arises in many practical domains, including  biology and architecture. Existing techniques assume the vectors to have the same domain or to be jointly distributed, and hence are not applicable. To address this, we propose Canonical Divergence Analysis (\ourmethod). We introduce three instantiations, each of which permits practical implementation. Extensive empirical evaluation shows the potential of our method.
\end{abstract}

\section{Introduction} \label{sec:intro}

In many application domains we want to  analyse the relation between data sets (views) of different sizes, dimensionality, and often also without sample-to-sample correspondence (i.e.\ no joint distribution). For instance, in metabolomics the goal is to study (dis)similarities of metabolic profiles collected from different populations, possibly different species, e.g.\ humans and mice~\cite{oresic:biomedical,tripathi:matching}. The results enable combining profiles from different organisms for joint analysis, leading to a more comprehensive view of biological properties. For energy efficient architecture, on the other hand, it is useful to know the impact of various energy and climate indicators over different buildings~\cite{wagner:building}. As the data sets may be on different domains, e.g.\ different sets of indicators with different cardinality recorded by heterogeneous technologies, to facilitate this it is necessary to find mappings that reveal similarities. In general, this problem also appears in e.g.\ comparing subspace clusters~\cite{tatti:apple}, finding structural similarities between different communities in a social network~\cite{traud:community}, and matching schemas between (legacy) databases~\cite{zhang:automatic}.

Formally, the problem can be cast as follow. Consider two random vectors, i.e.\ two sets of random variables. The first vector $\Xb \in \mathbb{R}^\DimX$ has $\SizeX$ realizations $\xb_1, \ldots, \xb_\SizeX$. The second $\Yb \in \mathbb{R}^\DimY$ has $\SizeY$ realizations $\yb_1, \ldots, \yb_\SizeY$. In general, $\DimX \neq \DimY$ and $\SizeX \neq \SizeY$. That is, $\Xb$ and $\Yb$ may differ on number of attributes \emph{as well as} on number of realizations. Further, there is no sample-to-sample correspondence between them, i.e.\ they do not have a joint distribution. Our goal is to find mappings of $\Xb$ and $\Yb$ maximizing their similarity, or equivalently minimizing their divergence.

Surprisingly, existing work does not address this general problem. 
Multi-view learning~\cite{sridharan:multiview}, e.g.\ canonical correlation analysis~\cite{hotelling:cca}, requires that $\Xb$ and $\Yb$ are jointly distributed; that is, they must have a sample-to-sample correspondence. 
Transfer learning techniques~\cite{pan:transfer} on the other hand in general require that $\Xb$ and $\Yb$ are from the same domain, i.e.\ $\DimX = \DimY$. Sample-to-sample correspondence analysis~\cite{tripathi:matching} assumes that each realization of $\Xb$ is paired with exactly one realization of $\Yb$. Thus, we can use neither to analyze data from e.g.\ different species or buildings. 

To allow for meaningful analysis of data sets of different sizes, dimensionality, and possibly without a joint distribution, we propose Canonical Divergence Analysis (\ourmethod). We instantiate it with three different divergence measures, leading to practical solutions. Further, we introduce two alternate formulations of \ourmethod, each with specific strengths; in particular, one for global optimization and one for fast optimization. Extensive experiments on a wide range of tasks demonstrate the potential of our approach.


\section{Formulation} \label{sec:formu}

We denote $r = \min\{\SizeX, \SizeY\}$. We also write $\vecu$ for a $\DimX \times 1$ vector and $\vecv$ for a $\DimY \times 1$ vector. We define \ourmethod as follow.

\begin{definition} \label{def:cda}
Canonical Divergence Analysis (\ourmethod) iteratively solves for $\{(\vecu^i, \vecv^i)\}_{i=1}^r$ where $\vecu^i \in \mathbb{R}^\DimX$ and $\vecv^i \in \mathbb{R}^\DimY$ at the $\mathit{i^{th}}$ iteration are the solutions of
$$\arg\min_{\substack{\vecu^T \vecu = \vecv^T \vecv = 1\\\vecu^T \vecu^j = 0, \forall j \in [1, i-1]\\\vecv^T \vecv^j = 0, \forall j \in [1, i-1]}} \diff\left(p(\vecu^T \Xb), q(\beta\vecv^T \Yb)\right)$$
with $p(.)$ and $q(.)$ being pdfs, $\diff$ being a divergence measure of pdfs, and $\beta \neq 0$ being the scaling factor to bring $\vecu^T \Xb$ and $\vecv^T \Yb$ to the same domain.
\end{definition}

There are some points to note. First, the \emph{orthonormal constraints} here prevent the canonical vectors from \emph{degenerating}. In addition, the orthogonal constraints, i.e.\ the second and third constraints, are not applied when solving for $(\vecu^1, \vecv^1)$; they are used when learning $\{(\vecu^i, \vecv^i)\}_{i=2}^r$.

Second, one could compare $\vecu^T \Xb$ and $\vecv^T \Yb$ directly. They may however not be on the same domain, potentially causing biases. For fair comparison, we use the scaling factor $\beta$, which is also learned.

Third, in \ourmethod we quantify the (dis)similarity of the transformed data $\vecu^T \Xb$ and $\beta\vecv^T \Yb$ through the divergence of their pdfs. This has intuition from statistics. In particular, the Mallows distance~\cite{rachev:wasser} of two univariate/multivariate random variables with no joint distribution is equal to the minimum of their expected difference -- defined by a given metric -- taken over \emph{all} possible joint distributions that preserve their marginal pdfs. Mallows distance is more commonly known as the Earth Mover's Distance~\cite{levina:mallows}, a divergence measure. In other words, Mallows distance tells us that analyzing the relationship between two random variables, taking into account all permissible joint distributions, reduces to computing the divergence of their distributions.

Fourth, \ourmethod has connections to the normalized information distance (\nid)~\cite{cilibrasi:comp} from information theory. Loosely speaking, \nid is based on the minimal Kolmogorov complexity of expressing $\Xb$ given $\Yb$, or vice versa. That is, \nid considers the length of the shortest algorithm that can generate one data set given the other. Mapping to \ourmethod, this generation is done through finding $\vecu$, $\vecv$, and $\beta$; the complexity is expressed by $\diff\left(p(\vecu^T \Xb), q(\beta\vecv^T \Yb)\right)$.

Furthermore, \ourmethod could be considered as a generalization of transfer component analysis (\tca)~\cite{pan:tca}. In short, \tca assumes that $\Xb$ and $\Yb$ are from the \textit{same} domain, i.e.\ $\DimX = \DimY$. It learns a \textit{common} mapping $\phi$ that minimizes $\diff(p(\phi(\Xb)), q(\phi(\Xb)))$. \ourmethod in turn allows $\DimX \neq \DimY$ and hence seeks for two different mappings. In other words, \ourmethod considers a more general problem setting.

Finally, \ourmethod focuses on linear mappings, which facilitates interpretation and post analysis~\cite{chang:cca,kara:cda}. It however does not limit \ourmethod to linear relations; the complexity of the latent similarity discovered is determined by $\diff$. That is, if $\diff$ is non-linear, \ourmethod will be able to discover non-linear relationships. Further non-linear mappings can be achieved by searching for kernel transformations~\cite{hardoon:kcca} or performing deep learning~\cite{andrew:cca}. As this is the first work on \ourmethod, we hence focus on the problem formulation and how to instantiate it for practical analysis -- we postpone extending \ourmethod to non-linear mappings to future work, but do note that our experiments show that with our current formulation far more complex relations than linear can be discovered.

Having introduced \ourmethod, next we review related work.

\section{Related Work} \label{sec:rl}

Multi-view learning~\cite{sridharan:multiview} searches for transformations of $\Xb$ and $\Yb$ of different domains. For instance, \cca~\cite{hotelling:cca} and its extensions~\cite{andrew:cca,chang:cca,hardoon:kcca,yin:cca} study the relationships between $\Xb$ and $\Yb$ by searching for mappings maximizing their correlation. For correlation analysis, the joint distribution of $\Xb$ and $\Yb$ must exist. This means that $\SizeX = \SizeY$ and every realization of $\Xb$ corresponds to a unique realization of $\Yb$ (and vice versa). Therewith \cca and multi-view learning in general address a different, more specific problem.

\ourmethod is related to multi-target regression, e.g.~\cite{aho:regression}, in the sense that one can perceive $\Xb$ as the independent variables and $\Yb$ as the targets. This setting however also requires $\Xb$ and $\Yb$ to have a sample-to-sample correspondence.

Sample-to-sample correspondence analysis~\cite{tripathi:matching} assumes that each realization of $\Xb$ is paired with exactly one realization of $\Yb$, but not necessarily the reverse. It first searches for mappings of $\Xb$ and $\Yb$ to bring them to a common domain. Then, it identifies a permutation of the realizations of $\Yb$ after mapping that minimizes the difference to the mapped realizations of $\Xb$. By assuming sample correspondence, it is not suited to analysing data taken from e.g.\ different species or different buildings. \ourmethod in contrast does not make this assumption and hence is more flexible.

In transfer learning~\cite{pan:transfer}, $\Xb$ and $\Yb$ can stand for the variables in the source and target domains. The goal is to learn transformations of $\Xb$ and $\Yb$ that are then used to boost performance of a learning task, e.g.\ classification or clustering, in the target domain. For instance, transductive transfer learning~\cite{taha:transfer} identifies mappings minimizing reconstruction error to the respective original space and classification error, i.e.\ its mappings is tailored to classification. Transfer component analysis~\cite{pan:tca}, assuming that $\Xb$ and $\Yb$ are from the same domain, learns a common mapping where distributions of the mapped $\Xb$ and $\Yb$ are similar. Likewise, self-taught learning~\cite{raina:stl}, structural correspondence learning~\cite{blitzer:scl}, and domain adaptation with coupled subspaces~\cite{blitzer:domain} assume that $\Xb$ and $\Yb$ are from the same domain. Self-taught clustering~\cite{dai:stl} searches for mappings of $\Xb$ and $\Yb$ that bring them to the same domain with the aim to help clustering in the target domain. It however works on discrete data and assumes that a common feature space of $\Xb$ and $\Yb$ is given. A comprehensive review of transfer learning literature can be found in~\cite{pan:transfer}. \ourmethod on the other hand is task-independent and works directly with real-valued data. Given $\Xb$ and $\Yb$ from \textit{different} domains and \textit{without} prior knowledge about their common space, we learn mappings that transform them to the same domain to facilitate joint analysis. 

Next, we solve \ourmethod by introducing different instantiations of divergence measure $\diff$. In Section~\ref{sec:beta} we discuss how to learn $\beta$ during the optimization process. In addition, in Section~\ref{sec:var} we propose two alternative ways to formulate the \ourmethod problem which enrich our options towards tackling it. Finally, we discuss our choice of the optimizer in Section~\ref{sec:opt}.

\section{Divergence Measure $\diff$} \label{sec:diff}

With $\diff$ we quantify the divergence between two univariate pdfs $p(\vecu^T \Xb)$ and $q(\beta\vecv^T \Yb)$. If we want to optimize $\diff$ it helps if it is differentiable. We therefore focus on three such divergence measures. The first is an extension of Mallows distance~\cite{rachev:wasser}. The second and third are based on kernel density estimation (\kde)~\cite{silverman:density}. Note that we are not constrained by these three measures; any differentiable divergence measure, e.g.\ Maximum Mean Discrepancy~\cite{borgwardt:mmd}, can be straightforwardly plugged into our solution.

Before going into the details, we introduce some notation. In general, it only makes sense to compare $p(\vecu^T \Xb)$ and $q(\beta\vecv^T \Yb)$ if they are defined on the same univariate random variable. We denote such a variable as $Z$. Hence, $p(\vecu^T \Xb)$ is equivalent to $p(Z)$ and $q(\beta\vecv^T \Yb)$ is equivalent to $q(Z)$. 
Under either notation we write the i.i.d.\ realizations of $p(.)$ as $\x_1, \ldots, \x_\SizeX$ where $\x_i = \vecu^T \xb_i$. Analogously, we write the i.i.d.\ realizations of $q(.)$ as $\y_1, \ldots, \y_\SizeY$ where $\y_j = \beta\vecv^T \yb_j$.

\subsection{Extended Mallows Distance}
\label{sec:instant:mallows}

For our first measure we extend the well-known Mallows distance~\cite{rachev:wasser}. For exposition, we denote $\vecu^T \Xb$ as $\X$ and $\beta\vecv^T \Yb$ as $\Y$. Our goal is to measure $\diff\left(p(\X), q(\Y)\right)$. Let $\mathcal{F}$ be the set of joint pdfs $f(\X, \Y)$ of $\X$ and $\Y$ such that the marginal distributions of $f$ on $\X$ and $\Y$ are $p(\X)$ and $q(\Y)$, respectively. That is, $\int_{\dom(\Y)} f(\x, \y) d\y = p(\x) \quad \forall \x \in \dom(\X)$, and $\int_{\dom(\X)} f(\x, \y) d\x = q(\y) \quad \forall \y \in \dom(\Y)$.

Given $t \geq 1$, the $t$-th Mallows distance between $p(\X)$ and $q(\Y)$ is~\cite{levina:mallows}
$$\mathit{MA}_t(p(\X), q(\Y)) = \left(\min\limits_{f \in \mathcal{F}} \int |\x - \y|^t f(\x, \y) d\x d\y\right)^{1/t}$$
If $p(\X)$ and $q(\Y)$ have the same number of realizations ($\SizeX = \SizeY$), we have
$\mathit{Mallows}_t(p(\X), q(\Y)) = \left(\frac{1}{\SizeX} \min_{\{j_1, \ldots, j_\SizeX\}} \sum_{i=1}^\SizeX |\x_i - \y_{j_i}|^t\right)^{1/t}$
where $\{j_1, \ldots, j_\SizeX\}$ is a permutation of $\{1, \ldots, \SizeX\}$.
If $\SizeX \neq \SizeY$, a possible solution is to replicate each realization so that both distributions have $\SizeX \times \SizeY$ realizations each. This practice yields the same Mallows distance as it preserves the empirical distributions.

Calculating Mallows distance on empirical data requires us to select the best permutation, which is expensive~\cite{levina:mallows}. To avoid this, we propose a modified version of Mallows distance. In particular, we set
$$\diff(p(\X), q(\Y)) = \left(\sum\limits_{f \in \mathcal{F}} \int |\x - \y|^t f(\x, \y) d\x d\y\right)^{1/t}$$
The difference between this version of $\diff$ and vanilla Mallows distance is that the latter takes the minimum while we consider the \emph{sum} over all $f \in \mathcal{F}$. This difference is reminiscent to the relationship between Cram\'er-von Mises and Kolmogorov-Smirnov tests. In particular, the former computes the integral of the (squared) difference between two cumulative distribution functions (cdfs) over the domain of the underlying random variable. The latter in turn takes the maximum (supremum) difference between two cdfs. We prove the following result.

\begin{lemma} \label{lem:mallows}
Empirically, minimizing $\diff(p(\X), q(\Y))$ is equivalent to minimizing $\left(\sum_{i=1}^\SizeX \sum_{j=1}^\SizeY |\x_i - \y_j|^t\right)^{1/t}$.
\end{lemma}

\proofApx

For simplicity, we use the function $\sum_{i=1}^\SizeX \sum_{j=1}^\SizeY |\x_i - \y_j|^t$ for optimization. In this paper, we set $t = 2$ and have
$$\diff(p(\X), q(\Y)) = \sum_{i=1}^\SizeX \sum_{j=1}^\SizeY (\x_i - \y_j)^2.$$
That is, we resemble $\diff(p(\X), q(\Y))$ as a least-squares cost function.  We will refer to \ourmethod using the extended Mallows distance as \mcda.

Interestingly, \mcda is closely related to linear \cca following the below lemma by Golub and Zha~\cite{golub:cca}.

\begin{lemma} \label{lem:cca}
Assume that $\Xb$ and $\Yb$ are whitened. Let $(\vecu^i, \vecv^j)$ be the $\mathit{i^{th}}$ canonical pair of linear \cca. Then
$$(\vecu^i, \vecv^i) = \arg\min_{\substack{\vecu^T \vecu = \vecv^T \vecv = 1\\\vecu^T \vecu^j = 0, j \in [1, i-1]\\\vecv^T \vecv^j = 0, j \in [1, i-1]}} ||\vecu^T \Xb - \vecv^T \Yb||^2.$$
\end{lemma}

Note that in the context of \cca, $\SizeX = \SizeY$ and each row of $\Xb$ is connected to exactly one row of $\Yb$. If we let $\x_i = \vecu^T \xb_i$ and $\y_i = \vecv^T \yb_i$ then linear \cca is equivalent to finding $(\vecu, \vecv)$ minimizing $\sum_{i=1}^\SizeX (\x_i - \y_i)^2$. Therefore, \mcda could be seen as a generalization of linear \cca when $\SizeX$ and $\SizeY$ may differ from each other, and there is no sample-to-sample correspondence between $\Xb$ and $\Yb$.

\subsection{Quadratic Measure}
\label{sec:instant:qr}

As the pdfs under consideration are univariate, \kde is applicable and reliable for density estimation. The \kde of $p(Z)$ is defined as $\widehat{p}(z) = \frac{1}{\SizeX \cdot \sigma_\Xb} \sum_{i=1}^\SizeX \kappa\left(\frac{z-\x_i}{\sigma_\Xb}\right) = \frac{1}{\SizeX} \sum_{i=1}^\SizeX \kappa_{\sigma_\Xb}(z-\x_i)$
where $\kappa(\cdot)$ is a symmetric but not necessarily positive kernel function that integrates to one, $\sigma_\Xb > 0$ is a smoothing parameter called the bandwidth, and $\kappa_{\sigma_\Xb}(z) = \frac{1}{\sigma_\Xb} \kappa(\frac{z}{\sigma_\Xb})$. Likewise, the \kde of $q(Z)$ is $\widehat{q}(z) = \frac{1}{\SizeY \cdot \sigma_\Yb} \sum_{j=1}^\SizeY \tau\left(\frac{z-\y_j}{\sigma_\Yb}\right) = \frac{1}{\SizeY} \sum_{j=1}^\SizeY \tau_{\sigma_\Yb}(z-\y_j)$ with $\tau(\cdot)$, $\sigma_\Yb$, and $\tau_{\sigma_\Yb}(\cdot)$ similarly defined.

We next instantiate $\diff(p(Z), q(Z))$ using one of the quadratic measures~\cite{rao:measure}. In particular, we set
$$\diff(p(Z), q(Z)) = \int \left(p(z) - q(z)\right)^2 dz$$
where $p(Z)$ and $q(Z)$ are approximated by \kde. Using \kde, empirically $\diff(p(Z), q(Z)) = $
\begin{align*}
& \frac{1}{\SizeX^2} \sum_{i=1}^\SizeX \sum_{j=1}^\SizeX \kappa_{\sigma_\Xb}(\x_i-\x_j) + \frac{1}{\SizeY^2} \sum_{i=1}^\SizeY \sum_{j=1}^\SizeY \tau_{\sigma_\Yb}(\y_i-\y_j) \\
& - \frac{1}{\SizeX \cdot \SizeY} \sum_{i=1}^\SizeX \sum_{j=1}^\SizeY \tau_{\sigma_\Yb}(\x_i-\y_j) - \frac{1}{\SizeX \cdot \SizeY} \sum_{i=1}^\SizeX \sum_{j=1}^\SizeY \kappa_{\sigma_\Xb}(\x_i-\y_j)
\end{align*}

In practice, any kernel functions can be used to instantiate $\kappa(\cdot)$ and $\tau(\cdot)$. In this work, we use the popular Gaussian kernel, i.e.\ we set $\kappa(z) = \tau(z) = \frac{1}{\sqrt{2\pi}} e^{-\frac{z^2}{2}}$. Following~\cite{chang:cca,gretton:hsic}, we set $\sigma_\Xb = \mathit{median}\{||\xb_i - \xb_j||: 1 \leq i, j \leq \SizeX\}$ and $\sigma_\Yb = \mathit{median}\{||\beta(\yb_i - \yb_j)||: 1 \leq i, j \leq \SizeY\}$. Note that more advanced empirical estimation of the above quadratic measure is available, e.g.~\cite{krishna:l22}. We stick to the current estimation due to its simplicity and its wide adoption in practice.

We will refer to \ourmethod using this quadratic measure as \qcda.

\subsection{Pearson Divergence}
\label{sec:instant:js}

As the third measure, we instantiate $\diff(p(Z), q(Z))$ by relative Pearson divergence~\cite{pearson}, or \pe divergence for short. That is, we set $\diff(p(Z), q(Z)) =$
\begin{align*}
& \int \left(\frac{1}{2} p(z) + \frac{1}{2} q(z)\right) \left(\frac{p(z)}{\frac{1}{2} p(z) + \frac{1}{2} q(z)} - 1\right)^2 dz\\
& + \int \left(\frac{1}{2} p(z) + \frac{1}{2} q(z)\right) \left(\frac{q(z)}{\frac{1}{2} p(z) + \frac{1}{2} q(z)} - 1\right)^2 dz.
\end{align*}
We denote the first and second terms of $\diff(p(Z), q(Z))$ as $\pe(p(Z)\; ||\; q(Z))$ and $\pe(q(Z)\; ||\; p(Z))$, respectively. Next, we present the computation of $\pe(p(Z)\; ||\; q(Z))$. The computation of $\pe(q(Z)\; ||\; p(Z))$ follows similarly.

One could compute $\pe(p(Z)\; ||\; q(Z))$ by estimating $p(Z)$ and $q(Z)$ using \kde. However, as \pe divergence is involved in ratio of pdfs, this would magnify the estimation error. To avoid this issue, we follow~\cite{liu:ulsif} and model the ratio $\frac{p(z)}{\frac{1}{2} p(z) + \frac{1}{2} q(z)}$ by the kernel model
$g(z;\; \theta) = \sum_{i=1}^d \theta_i \omega(z, x'_i)$
where $\theta = (\theta_1, \ldots, \theta_d)$ are parameters to be learned from empirical data, $\{x'_1, \ldots, x'_d\} \subseteq \{\x_1, \ldots, \x_\SizeX\}$, and $\omega(z, x'_i)$ is the kernel basis function. The kernel centers $\{x'_1, \ldots, x'_d\}$ are drawn randomly from $\{\x_1, \ldots, \x_\SizeX\}$ where $d = \min(200, \SizeX)$ to increase efficiency~\cite{kara:cda}. In addition, we employ the widely used Gaussian RBF kernel $\omega(z, x'_i) = e^{-\frac{(z - x'_i)^2}{2\sigma_\Xb^2}}$. The bandwidth $\sigma_\Xb$ is set similarly to the case of the quadratic measure.

Computing $\theta$ requires us to minimize a squared loss objective function. Fortunately, this sub-problem has an analytical solution~\cite{liu:ulsif}. In particular, we have $\widehat{\theta} = (\E + \lambda \I_d)^{-1} \e$ where $\E$ is a $d \times d$ matrix with $\E_{i,j} = \frac{1}{2 \SizeX} \sum_{r=1}^\SizeX \omega(x_r, x'_i) \omega(x_r, x'_j) + \frac{1}{2 \SizeY} \sum_{r=1}^\SizeY \omega(y_r, x'_i) \omega(x_r, x'_j)$, $\e$ is a $d$ dimensional vector with $\e_i = \frac{1}{\SizeX} \sum_{j=1}^\SizeX \omega(x_j, x'_i)$, $\I_d$ is the $d \times d$ identity matrix, and $\lambda \geq 0$ is a regularization parameter chosen by cross-validation.

Therefore, we obtain $\widehat{g}(z) = \sum_{i=1}^d \widehat{\theta}_i \omega(z, x'_i)$. Empirically, $\pe(p(Z)\; ||\; q(Z)) =$
$$-\frac{1}{4 \SizeX} \sum_{i=1}^\SizeX \widehat{g}(x_i)^2 - \frac{1}{4 \SizeY} \sum_{j=1}^\SizeY \widehat{g}(y_j)^2 + \frac{1}{\SizeX} \sum_{i=1}^\SizeX \widehat{g}(x_i) - \frac{1}{2}.$$
We will refer to \ourmethod with \pe divergence as \pecda.

\section{Setting scaling factor $\beta$} \label{sec:beta}

To meaningfully compute the divergence between $\vecu^T \Xb$ and $\vecv^T \Yb$ we need to bring them to the same domain. We do so with scaling factor $\beta$. Obviously, we can optimize $\beta$ as part of the \ourmethod optimization problem. Alternatively, reducing the number of variables to optimize, we propose to directly set $\beta$ at each step of the optimization process, e.g.\ at each gradient descent of $\vecu$ and $\vecv$.

As standard in data analysis, before performing \ourmethod we rescale all attributes $\Xb_1, \ldots, \Xb_\DimX$ of $\Xb$ and $\Yb_1, \ldots, \Yb_\DimY$ of $\Yb$ to the same domain. W.l.o.g., we assume this domain to be $[0, 1]$. Let $\bar{\DimX}$ be the number of non-zero elements of $\vecu$ and $\bar{\DimY}$ be the number of non-zero elements of $\vecv$.

\begin{lemma} \label{lem:beta}
If $\beta = \sqrt{\frac{\bar{\DimX}}{\bar{\DimY}}}$ then $\vecu^T \Xb$ and $\beta \vecv^T \Yb$ have the same domain, which is $[-\sqrt{\bar{\DimX}}, \sqrt{\bar{\DimX}}]$.
\end{lemma}

\proofApx

This implies that we can gain efficiency by setting $\beta$ to $\sqrt{\frac{\bar{\DimX}}{\bar{\DimY}}}$. Though straightforward,  we prove that it is robust to noisy dimensions, as follows.

\begin{lemma} \label{lem:beta2}
Let $(\vecu, \vecv)$ be a pair of canonical vectors that \ourmethod discovers on $\Xb$ and $\Yb$, i.e.\ $\diff(p(\vecu^T \Xb), q(\beta\vecv^T \Yb))$ is small. Given an integer $c > 0$, let $\Xb' \in \mathbb{R}^{\DimX + c}$ be a random vector created by adding (noisy) attributes $\Xb_{\DimX + 1}, \ldots, \Xb_{\DimX + c}$ to $\Xb$. Given an integer $d > 0$, let $\Yb' \in \mathbb{R}^{\DimY + d}$ be a random vector created by adding (noisy) attributes $\Yb_{\DimY + 1}, \ldots, \Yb_{\DimY + d}$ to $\Yb$. It holds that $(\vecu, \vecv)$ are identifiable when solving \ourmethod on $\Xb'$ and $\Yb'$.
\end{lemma}

\proofApx

We empirically evaluate this property by testing with data sets that include noisy attributes (see Section~\ref{sec:retrieve} and Appendix~\ref{sec:noise}). We also compare our setting of $\beta$ to optimizing $\beta$; the results (see Appendix~\ref{sec:optimizing}) show that there is no clear winner while our method is more efficient.

\section{Alternate \ourmethod Formulations} \label{sec:var}

With our \ourmethod formulation, we achieve three instantiations that lead to practical solutions. This formulation however is not unique. In this section, we introduce two alternatives. In particular, instead of iteratively learning pairs of canonical vectors, we look at finding multiple pairs simultaneously, which may be good for avoiding local optima. As the second alternative, we replace the hard unit norm constraints $\vecu^T \vecu = \vecv^T \vecv = 1$ by soft reconstruction costs, and hence enable the application of fast unconstrained optimizers. The details are as follow.

\subsection{Multi-dimensional \ourmethod}

In principle, the first alternative is similar to the original formulation, except for that it searches for all pairs of vectors simultaneously instead of iteratively. This may be good for avoiding local optima. Formally, we have:

\begin{definition} \label{def:multicda}
Multi-dimensional \ourmethod (\Mcda) solves for $\U = (\vecu^1, \ldots, \vecu^r) \in \mathbb{R}^{\DimX \times r}$ and $\V = (\vecv^1, \ldots, \vecv^r) \in \mathbb{R}^{\DimY \times r}$ where $(\U, \V)$ is the solution of
$$\arg\min_{\U^T \U = \V^T \V = \mathbf{I}_r} \diff\left(p(\U^T \Xb), q(\mathbf{\Gamma} \V^T \Yb)\right)$$
with $p(.)$ and $q(.)$ being pdfs, $\diff$ being a divergence measure of pdfs, and $\mathbf{\Gamma} = \mathit{diag}(\beta_1, \ldots, \beta_r)$ being the scaling matrix to bring $(\vecu^i)^T \Xb$ and $(\vecv^i)^T \Yb$ to the same domain.
\end{definition}

To solve \Mcda, we need $\diff$ which is suited to multivariate settings and whose analytical form is easy to compute and differentiable. Note that multiplying univariate $\diff$ terms does not work as this requires dimensions in projected spaces to be statistically independent. Of the measures we introduced in Section~\ref{sec:diff}, \pe divergence meets this requirement and hence we use it for solving \Mcda.

\subsection{Reconstruction \ourmethod}

The unit norm constraints of the original \ourmethod problem may lead to slow optimization. To boost efficiency, we relax them using reconstruction costs~\cite{le:ica,bengio:deep,olshausen:sparse} and hence enable the application of fast unconstrained optimizers. We hence define the second alternative formulation as follow.

\begin{definition} \label{def:recda}
Reconstruction \ourmethod (\Rcda) iteratively solves for $\{(\vecu^i, \vecv^i)\}_{i=1}^r$ where $\vecu^i \in \mathbb{R}^\DimX$ and $\vecv^i \in \mathbb{R}^\DimY$ at the $\mathit{i^{th}}$ iteration are the solutions of
\begin{align*}
& \arg\min_{\substack{\vecu^T \vecu^j = 0, \forall j \in [1, i-1]\\\vecv^T \vecv^j = 0, \forall j \in [1, i-1]}} \left(\frac{\lambda}{\SizeX} \sum_{i=1}^\SizeX ||\vecu \vecu^T \tilde{\xb_i} - \tilde{\xb_i}||^2 \right.\\
& \left. + \frac{\delta}{\SizeY} \sum_{j=1}^\SizeY ||\vecv \vecv^T \tilde{\yb_j} - \tilde{\yb_j}||^2 + \diff\left(p(\vecu^T \Xb), q(\beta\vecv^T \Yb)\right)\right)
\end{align*}
with $\tilde{\xb_i} = \xb_i - \mu(\Xb_i)$ being the centered version of $\xb_i$, $\tilde{\yb_j} = \yb_j - \mu(\Yb_j)$ being the centered version of $\yb_j$, $p(.)$ and $q(.)$ being pdfs, $\diff$ being a divergence measure of pdfs, $\lambda > 0$ and $\delta > 0$ being weights of reconstruction costs, and $\beta \neq 0$ being the scaling factor to bring $\vecu^T \Xb$ and $\vecv^T \Yb$ to the same domain.
\end{definition}

We define Multi-dimensional \Rcda (\MRcda) analogously and give the definition in Appendix~\ref{sec:mrcda}. We show that \Rcda and the original \ourmethod are related by proving two following properties of the reconstruction costs.

\begin{proposition} \label{prop:recda}
If $\Xb$ is whitened, $\frac{\lambda}{\SizeX} \sum_{i=1}^\SizeX ||\vecu \vecu^T \tilde{\xb_i} - \tilde{\xb_i}||^2$ is equivalent to $\lambda ||\vecu \vecu^T - \mathbf{I}_\DimX||_F^2$ where $\mathbf{I}_\DimX$ is the $\DimX \times \DimX$ identity matrix and $||.||_F$ denotes the Frobenius norm.
\end{proposition}

\proofApx

\begin{proposition} \label{prop:recda2}
$\lambda ||\vecu \vecu^T - \mathbf{I}_\DimX||_F^2$ is equivalent to $\lambda (\vecu^T \vecu - 1)^2$ up to an additive constant.
\end{proposition}

\proofApx

Based on Propositions~\ref{prop:recda} and~\ref{prop:recda2}, we prove the following.

\begin{lemma} \label{lem:recda}
If $\Xb$ and $\Yb$ are whitened, $\lambda \uparrow +\infty$, and $\delta \uparrow +\infty$, then 1) \Rcda is equivalent to \ourmethod and 2) \MRcda is equivalent to \Mcda.
\end{lemma}

\proofApx

With the the unit norm constraints relaxed, we can solve \Rcda and \MRcda with fast unconstrained solvers.

\section{Optimization} \label{sec:opt}

For the original \ourmethod formulation, we try with different optimizers; in particular, projected gradient descent~\cite{edelman:gradient}, sequential quadratic programming~\cite{nocedal:opt}, and natural gradient descent~\cite{amari:gradient}. Empirically, we find the last option to yield the best results. Therefore, we use it in our experiment.
The update rule of this optimizer is that
$\vecu \leftarrow \vecu \times \exp\left(-t \left(\vecu^T \nabla_\vecu f - \nabla_\vecu f^T \vecu \right)\right)$ and 
$\vecv \leftarrow \vecv \times \exp\left(-t \left(\vecv^T \nabla_\vecv f - \nabla_\vecv f^T \vecv \right)\right)$
where $\exp$ for a matrix denotes the \textit{matrix exponential} and $t > 0$ is the step size. We choose the optimal step size by Nelder-Mead method.

For \Mcda, we also use natural gradient descent.

\Rcda and \MRcda are unconstrained optimization problems. As optimizer, similar to~\cite{le:ica} we use L-BFGS and Conjugate Gradient and present the results with the former. In practice, we see that \cda methods using reconstruction costs are about 5 times faster than their respective counterpart. We provide more details in Appendix~\ref{sec:time}.

\section{Experiments} \label{sec:exp}

In the following, we represent the data of $\Xb$ as a $\SizeX \times \DimX$ matrix $\D_\Xb$ where the $i$-th row of $\D_\Xb$ corresponds to $\xb_i$. We define the $\SizeY \times \DimY$ matrix $\D_\Yb$ similarly. We empirically evaluate our methods on four tasks. First, using synthetic data sets we show that they are able to retrieve known relations. Second, using two data sets from UCI Repository~\cite{uci:2010} we show that they can be used in cross-domain regression and classification. Third, we use them to discover interesting relations between two real-world data sets taken from studies on architecture~\cite{schweiker:building,wagner:building}. Last, we plug our methods into \rescu~\cite{rescu} to discover non-redundant subspace clusters.

We compare to the original \cca~\cite{hotelling:cca} and \lscca~\cite{kara:cda}---a non-linear \cca method; \scl~\cite{blitzer:scl} and \dac~\cite{blitzer:scl} for transfer learning; \ssc~\cite{tripathi:matching} for sample-to-sample correspondence detection.
We implemented our methods in Matlab and provide the code in the supplementary material.

\subsection{Retrieving Canonical Vectors} \label{sec:retrieve}

First we evaluate whether our methods can recover associations between data sets where we know the ground truth. To do so, we generate pairs ($\D_\Xb$, $\D_\Yb$) with relationships between $\Xb$ and $\Yb$ embedded a priori. In particular, we first generate $\SizeX = 1000$ samples for $\Xb = (\Xb_1, \ldots, \Xb_\DimX)$ where $\DimX \geq 5$ and $\Xb_i \sim \mathcal{N}(0, 1)$. Then we generate $\SizeY$ samples for $\Yb = (\Yb_1, \ldots, \Yb_\DimY)$ with $\DimY \geq 5$ according to three types of relations.

\textit{Linear relations:}
\begin{flalign*}
& \textstyle\Yb_1 = \Xb_1 + 2\Xb_2 + \epsilon_1 & \Yb_2 = \Xb_3 + 2\Xb_4 + \epsilon_2\\
& \textstyle\Yb_3 = \Xb_5 + \epsilon_3 & \Yb_4 = \Xb_2 + \Xb_5 + \epsilon_4\\
& \textstyle\Yb_{5, \ldots, \DimY} \sim \mathcal{N}(0, 1) & \epsilon_{1, \ldots, 4} \sim \mathcal{N}(0, 0.5)
\end{flalign*}

\textit{Mixed relations:}
\begin{flalign*}
& \Yb_1 = \Xb_1^2 + 2\Xb_2 + \epsilon_1 & \Yb_2 = \Xb_3^3 + 2\Xb_4 + \epsilon_2\\
& \Yb_3 = \Xb_5 + \epsilon_3 & \Yb_4 = \Xb_2 + \Xb_5 + \epsilon_4\\
& \Yb_{5, \ldots, \DimY} \sim \mathcal{N}(0, 1) & \epsilon_{1, \ldots, 4} \sim \mathcal{N}(0, 0.5)
\end{flalign*}

\textit{Non-linear relations:}
\begin{flalign*}
& \Yb_1 = \Xb_1^2 + 2\Xb_2 + \epsilon_1 & \Yb_2 = \Xb_3^3 + 2\Xb_4 + \epsilon_2\\
& \Yb_3 = e^{\Xb_5} + \epsilon_3 & \Yb_4 = cos(\Xb_2 + \Xb_5) + \epsilon_4\\
& \Yb_{5, \ldots, \DimY} \sim \mathcal{N}(0, 1) & \epsilon_{1, \ldots, 3} \sim \mathcal{N}(0, 0.5),\; \epsilon_4 \sim \mathcal{N}(0, 0.1)
\end{flalign*}

We refer to the canonical vectors discovered by a method as $\U = (\vecu_1, \ldots, \vecu_r)$ and $\V = (\vecv_1, \ldots, \vecv_r)$ where $r = \min\{\DimX, \DimY\}$. That is, each column of $\U$ and $\V$ is a canonical vector. Let $\U_{\mathit{gr}}$ and $\V_{\mathit{gr}}$ be the respective ground truth. Similarly to~\cite{kara:cda}, we quantify performance as
\begin{align*}
\textstyle\frac{1}{2\sqrt{2r}} \left(||\U \U^T - \U_{\mathit{gr}} \U_{\mathit{gr}}^T||_{F} + ||\V \V^T - \V_{\mathit{gr}} \V_{\mathit{gr}}^T||_{F}\right)
\end{align*}
where $||.||_{F}$ stands for the Frobenius norm. Intuitively, the smaller the value the better the result.

We experiment with three settings. First, we generate $\D_\Xb$ and $\D_\Yb$ where $\SizeX = \SizeY$ and their row order is intact. Second, we keep $\SizeX = \SizeY$ but randomize the row order of $\D_\Xb$ and $\D_\Yb$. Third, we generate data sets where $\SizeX = \SizeY$; then we randomly remove $\rho \times \SizeY$ records of $\D_\Yb$ where $\rho \in \{0.05, 0.1, 0.15, 0.2\}$. In all settings, $\DimX = 7$ and $\DimY = 5$, i.e.\ $\Xb$ has two noisy dimensions and $\Yb$ has one noisy dimension. As \scl and \dac requires that $\Xb$ and $\Yb$ come from the same domain, to test them we skip the 2 noisy dimensions of $\Xb$. Regarding our methods, for succinctness we present the outcome on \qcda (normal \ourmethod with the quadratic measure), \mRcda (Reconstruction \ourmethod with the extended Mallows distance), and \peMRcda (Multi-dimensional Reconstruction \ourmethod with \pe divergence). For \mRcda and \peMRcda, similar to~\cite{le:ica} we set $\lambda = \delta = 0.5$.

We give the results in Table~\ref{tab:retrieve}. For the first two settings, each value is the average of 10 runs; standard deviation is small and hence skipped. For the last setting, each value is the average over all runs corresponding to all values of $\rho$ tested and 10 runs per value. Going over the results, we make the following observations.

\begin{table*}[tb]
\centering
\small
\begin{tabular}{llrrrrrrrr}
\toprule
& & \multicolumn{3}{l}{\textbf{\cda}} & & & & &\\
\cmidrule(r){3-5}
Setting & \textbf{Type} & \textbf{\qcda} & \textbf{\mRcda} & \textbf{\peMRcda} & \textbf{\cca} & \textbf{\lscca} & \textbf{\scl} & \textbf{\dac} & \textbf{\ssc}\\

\otoprule

\multirow{2}{*}{\emph{$\SizeX = \SizeY$}} & Linear & 0.31 & 0.30 & 0.31 & \textbf{0.26} & 0.45 & 0.38 & 0.38 & 0.30\\

\multirow{3}{*}{\emph{row order intact}} & Mixed & \textbf{0.34} & \textbf{0.34} & \textbf{0.35} & 0.41 & 0.45 & 0.48 & 0.51 & 0.46\\

& Non-linear & \textbf{0.36} & \textbf{0.35} & \textbf{0.35} & 0.50 & 0.48 & 0.52 & 0.57 & 0.54\\

\midrule
\midrule

\multirow{2}{*}{\emph{$\SizeX = \SizeY$}} & Linear & \textbf{0.31} & \textbf{0.30} & \textbf{0.31} & 0.48 & 0.53 & 0.38 & 0.38 & \textbf{0.31}\\

\multirow{3}{*}{\emph{row order randomized}} & Mixed & \textbf{0.34} & \textbf{0.34} & \textbf{0.35} & 0.62 & 0.55 & 0.48 & 0.51 & 0.47\\

& Non-linear & \textbf{0.36} & \textbf{0.35} & \textbf{0.35} & 0.69 & 0.57 & 0.52 & 0.57 & 0.54\\

\midrule
\midrule

\multirow{3}{*}{\emph{$\SizeX \neq \SizeY$}} & Linear & 0.33 & \textbf{0.31} & \textbf{0.30} & \emph{n/a} & \emph{n/a} & 0.44 & 0.41 & 0.35\\

& Mixed & \textbf{0.36} & \textbf{0.36} & 0.38 & \emph{n/a} & \emph{n/a} & 0.49 & 0.53 & 0.52\\

& Non-linear & \textbf{0.39} & \textbf{0.38} & \textbf{0.39} & \emph{n/a} & \emph{n/a} & 0.53 & 0.60 & 0.56\\

\bottomrule
\end{tabular}
\caption{[Lower is better] Frobenius norm errors for different settings. All values are obtained after 10 trials. Best values and comparable ones according to the Wilcoxon signed rank test at $\alpha = 5\%$ are in \textbf{bold}.} \label{tab:retrieve} 
\end{table*}

In the first setting, \cca is best on linear relations. \mRcda and \ssc come in second. On mixed and non-linear relations, \textit{all} of our methods give the best results. This is remarkable as they do not use the joint distribution of $\Xb$ and $\Yb$.

In the second setting where row order is randomized, as expected, the performance of \qcda, \mRcda, and \peMRcda is not impacted at all. So are transfer learning and sample-to-sample correspondence methods. \cca methods in turn yield not good results, stemming from the fact that they require row order to work properly.

In the third setting, \cca methods are not applicable. Our methods in contrast are relatively stable under the down-sampling of $\D_\Yb$; the increase in error stays below 12\%, even for $\rho = 20\%$. This corroborates that \ourmethod is suited for analyzing data with neither known association nor the same number of records.

The results so far show \mRcda to have good performance across different settings. Thus, for the remaining experiments we pick it for exposition.

\subsection{Cross-domain Regression and Classification} \label{sec:cross}

We aim at studying if the mappings identified by \mRcda are useful for cross-domain learning tasks, e.g.\ regression and classification. Cross-domain regression and classification are popular forms to assess transfer learning techniques, e.g.~\scl~\cite{blitzer:scl}, \dac~\cite{blitzer:scl}, and \tca~\cite{pan:tca}. As \ourmethod could be perceived as a generalization of \tca, it is interesting to study if the mappings identified by our methods, in particular \mRcda, are also useful for these tasks.

For cross-domain regression, we use the Bike data from UCI Repository. It consists of two data sets with the same attribute set, one recording bike rental statistics per hour ($\D_\Xb$), one recording per day ($\D_\Yb$). The target variable is the number of bikers. Assuming the target values in $\D_\Yb$ are not available, i.e.\ $\D_\Yb$ is unlabeled, the task is to predict the number of bikers of $\D_\Yb$ using $\D_\Xb$. Similar to the setup of Pan et al.~\cite{pan:tca}, we first use \mRcda to learn the mappings of $\Xb$ and $\Yb$ that bring them to the same latent space. We then apply Regularized Least Square Regression (RLSR), training in the latent space using labeled $\D_\Xb$ and some randomly selected records of $\D_\Yb$. Finally, we evaluate the performance using the remaining records of $\D_\Yb$. We use Average Error Distance as performance metric. Each result shown is the average of 10 runs.

For cross-domain classification, we use the Energy data described in~\cite{nguyen:4s}. $\D_\Xb$ and $\D_\Yb$ contain data collected from the same building in two non-overlapping periods, i.e.\ we consider the same domain. The class label is whether the given day is a week day or weekend. The testing setup is similar to above except that we use a SVM with linear kernel as classifier, and use Classification Error as metric.

\begin{figure}[t]
\centering
\subfigure[Average Error Distance vs.\ Number of Attributes]
{{\includegraphics[width=0.23\textwidth]{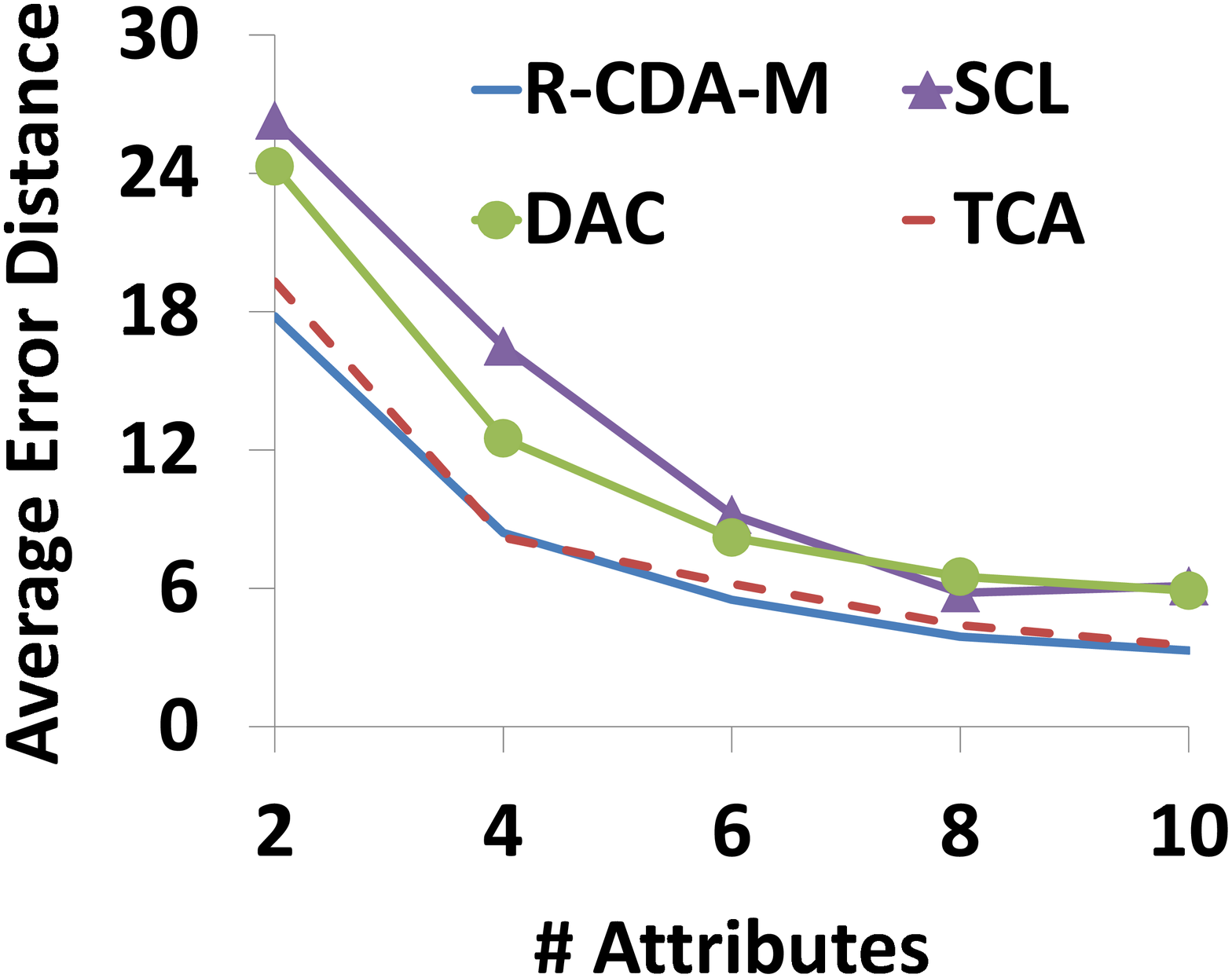}}\label{fig:aed_vs_dim}}
\subfigure[Classification Error vs.\ Number of Attributes]
{{\includegraphics[width=0.23\textwidth]{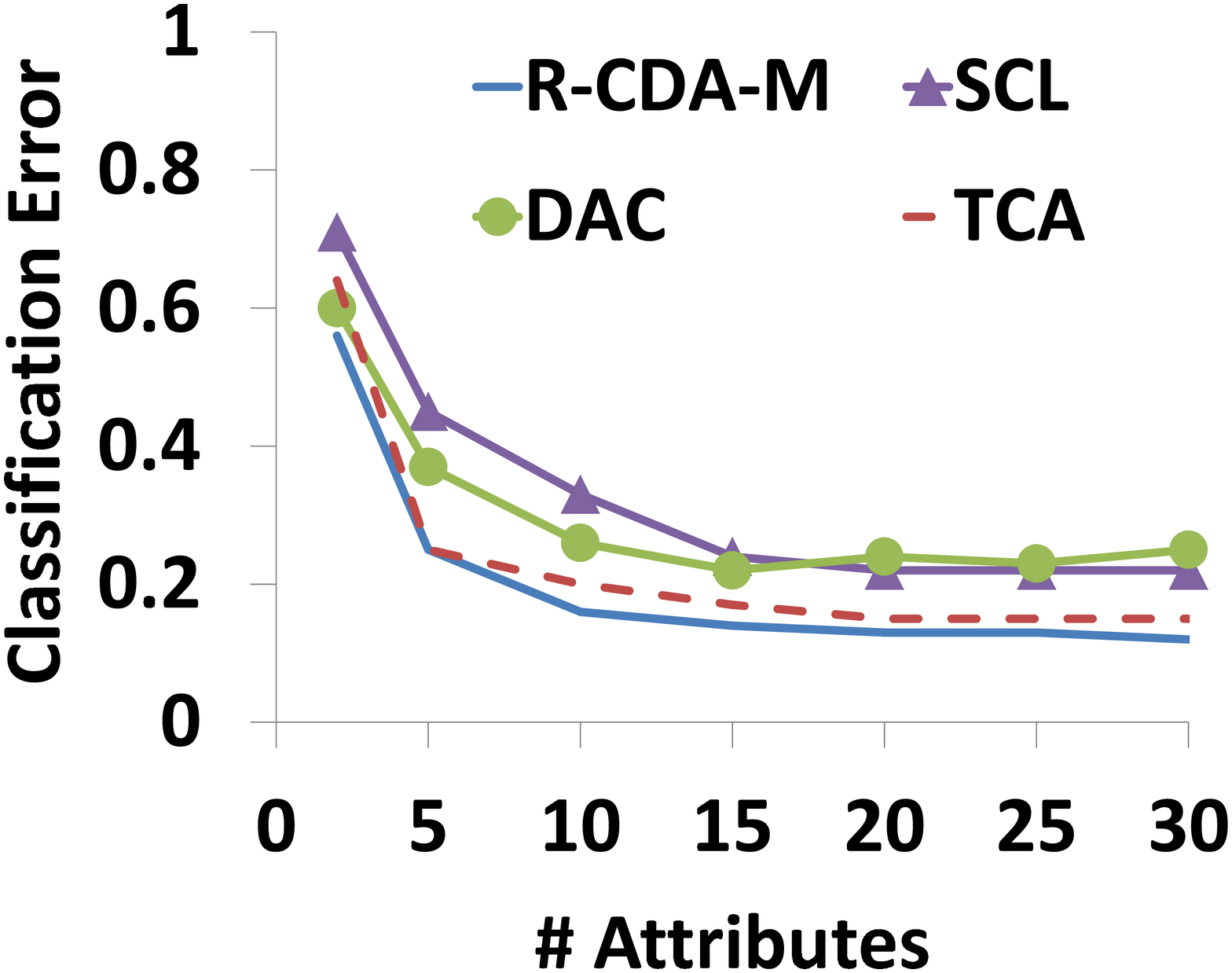}}\label{fig:error_vs_dim}}
\caption{[Lower is better] (left) Average Error Distance on the Bike data and (right) Classification Error on the Energy data against the number of attributes in the mapping space. For regression, we use Regularized Least Square Regression. For classification, we use SVM.}
\end{figure}

Note that, to enable comparison with \scl~\cite{blitzer:scl}, \dac~\cite{blitzer:scl}, and \tca~\cite{pan:tca}, in both experiments we chose $\D_\Xb$ and $\D_\Yb$ to have the same number of attributes -- this is not a requirement of \cda, however. The results are in Figures~\ref{fig:aed_vs_dim} and~\ref{fig:error_vs_dim}. We can see that \mRcda performs rather well, on par with \tca, suggesting it is a viable technique for cross-domain learning.

\subsection{Discovering Novel Relations} \label{sec:discovering}

We use \mRcda to discover relations between two data sets taken from two nearby office buildings with similar architectural characteristics. These data sets are from architectural studies~\cite{schweiker:building,wagner:building}. The first one ($\D_\Xb$) has 4421 records and 15 attributes; some of which are \textit{electricity consumption}, \textit{amount of water consumed}, and \textit{amount of heating}. The second one ($\D_\Yb$) has 3257 records and 20 attributes; some of which are \textit{electricity consumption}, \textit{indoor CO2 concentration}, and \textit{indoor air temperature}. Two data sets have 7 attributes in common. Further, they are collected in different time periods and hence there is no direct way to match their samples. 

Regarding the results, \mRcda identifies that attributes \textit{number of staff members} and \textit{percentage of building occupation} of $\Xb$ are related to attribute \textit{amount of heating} of $\Yb$. Taking into account the fact that two buildings are close to each other and have similar architectural characteristics, this relation is intuitively understandable. \mRcda also detects that attribute \textit{amount of water consumed} of $\Xb$ is related to attributes \textit{indoor CO$_2$ concentration} and \textit{indoor air temperature} of $\Yb$. Interestingly, a similar relation has recently been found in~\cite{nguyen:mac} for data from a \textit{single} building. That \mRcda discovers this relation for two different buildings could be attributed to their close proximity and architectural similarities.

\begin{figure}[t]
\centering
\subfigure[Average Error Distance vs.\ Number of Attributes]
{{\includegraphics[width=0.23\textwidth]{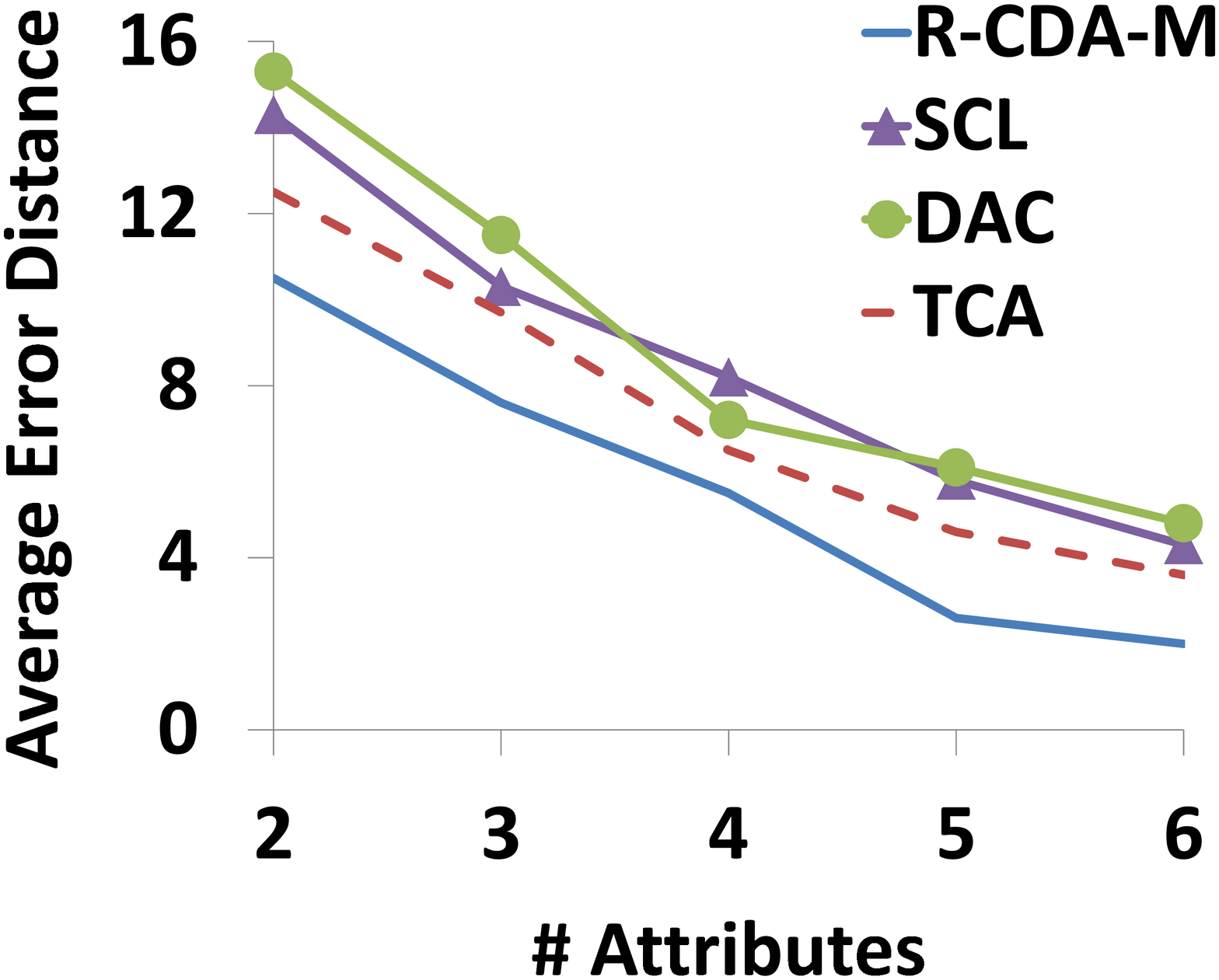}}\label{fig:clim_aed_vs_dim}}
\subfigure[Classification Error vs.\ Number of Attributes]
{{\includegraphics[width=0.23\textwidth]{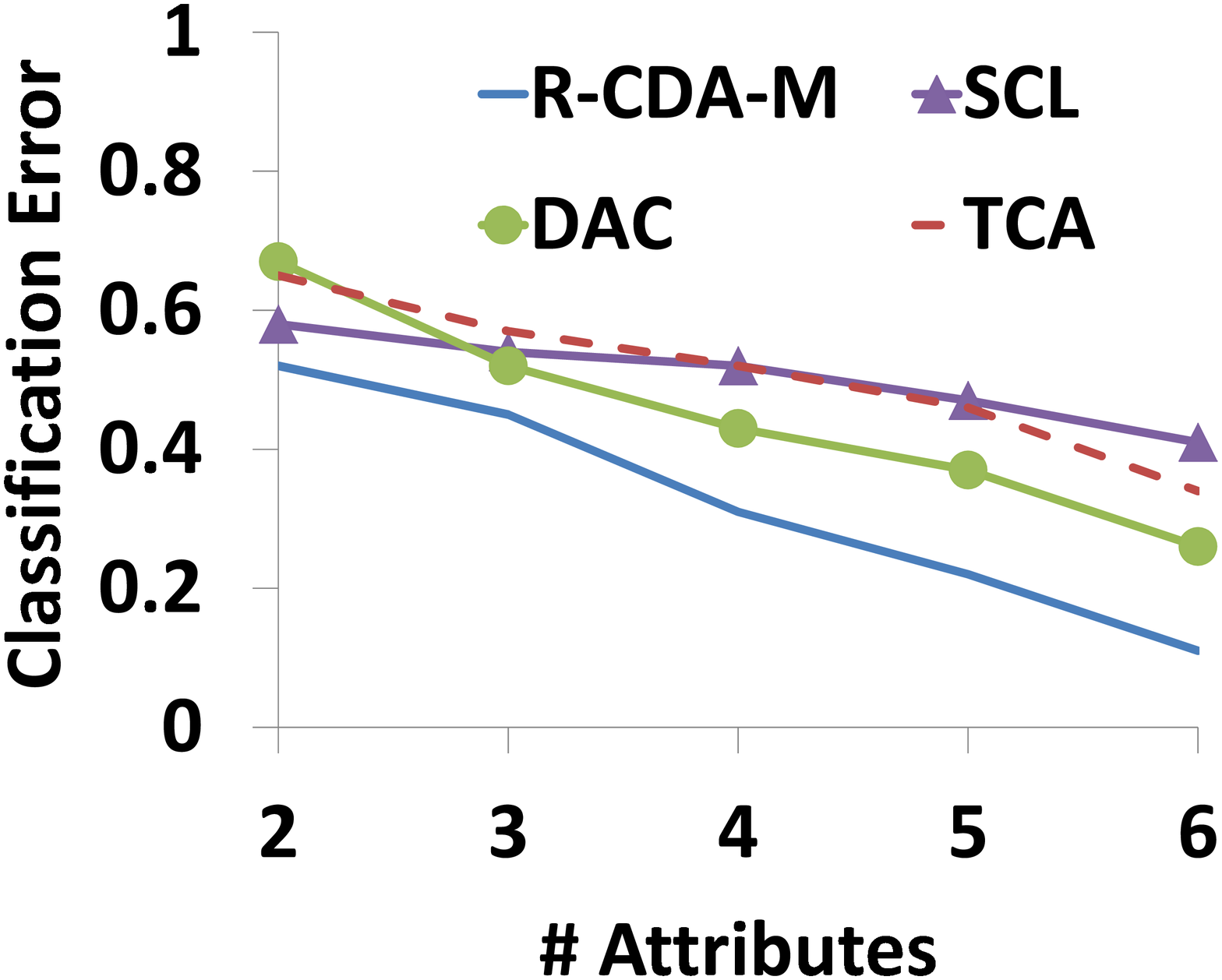}}\label{fig:clim_error_vs_dim}}
\caption{[Lower is better] (left) Average Error Distance and (right) Classification Error on the Building data against the number of attributes in the mapping space. For regression, we use Regularized Least Square Regression. For classification, we use SVM.}
\end{figure}

Besides, we repeat the cross-domain learning setup in Section~\ref{sec:cross}. For regression, the target variable is \textit{electricity consumption}. For classification, for each sample of both $\Xb$ and $\Yb$ we choose the weekday when it was recorded as its class label. To make \scl, \dac, and \tca applicable, when testing with them we use the common attributes only. The results are in Figures~\ref{fig:clim_aed_vs_dim} and~\ref{fig:clim_error_vs_dim}. \mRcda gives the best performance in both tasks. This could be attributed to the fact that \mRcda can use all attributes to learn the mappings. The other three methods require $\DimX = \DimY$ and hence do not take full advantage of the available information for joint analysis of $\Xb$ and $\Yb$.

\subsection{Non-redundant Subspace Clustering} \label{sec:clustering}

Next, we apply \ourmethod to compare subspace clusters. Subspace clustering aims at finding objects clustered in subsets, \emph{subspaces}, of all dimensions. There are two available approaches. One that finds subspace clusters that do not overlap in terms of objects~\cite{park:subspace} and one that does allow overlap in the objects~\cite{kriegel:survey}. We consider the latter setting. One of its major problems is redundancy in the output, as many clusters may be syntactically different but convey similar information. \rescu~\cite{rescu} is a recently proposed technique to alleviate the issue and improve post-analysis.

It works in an iterative manner. At each step, let $\csel$ be the set of subspace clusters already found and $\ccand$ be the set of candidate subspace clusters. The potential of each cluster $C \in \ccand$ w.r.t.\ $\csel$ is defined as $\potent(C, \csel) = \frac{|\cov(C) \setminus \bigcup_{C' \in \csel} \cov(C')|}{\cost(C)}$ where $\cov(C)$ is the set of objects covered by $C$ and $\cost(C)$ is the cost of mining $C$. \rescu then selects $C^{*} \in \ccand$ with largest potential. The process goes on until $\ccand$ becomes empty. That \rescu considers object counts only is simplistic.

To improve, we propose to set
$$\potent(C, \csel) = \frac{|\cov(C) \setminus \bigcup_{C' \in \csel} \cov(C')|}{\cost(C) \sum_{C' \in \csel} \dist(C, C')}$$
where $\dist(C, C')$ is the distance between $\D_C$ and $\D_{C'}$, respectively the data of $C$ and $C'$. Note that $\D_C$ and $\D_{C'}$ may have nonidentical sets of records and attributes.
We define $\dist$ as follow. For each \ourmethod method $\model$, we set
$$\textstyle\dist_\model(C, C') = \frac{1}{w} \sum_{i=1}^w \obj_\model(\vecu_i, \vecv_i)$$
where $w$ is the minimum of the dimensionality of $\D_C$ and $\D_{C'}$, $\obj_\model$ is the objective function of $\model$, and $(\vecu_i, \vecv_i)$ is the $\mathit{i^{th}}$ pair of canonical vectors identified by $\model$.

Therefore, we take into account also the similarity of $C$ to the clusters in $\csel$. We hypothesize that this would improve the clustering quality. Note that existing methods for comparing clusters assume that clusters are on the same space~\cite{romano:mi} and hence are not applicable here.

\begin{figure}[t]
\centering
\includegraphics[width=0.3\textwidth]{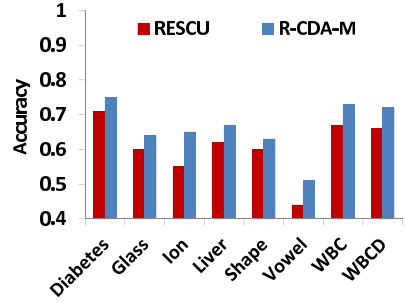}
\caption{[Higher is better] Clustering quality (F1 values) of original \rescu and \rescu with \mRcda on real-world data sets.} \label{fig:rescu}
\end{figure}

For evaluation, we use the same UCI data sets employed in~\cite{rescu}. We report average F1 scores, the harmonic mean of recall and precision, over 10 independent runs. The results are in Figure~\ref{fig:rescu}. We find that \mRcda consistently improves the clustering quality of the original \rescu for all data sets tested. This confirms both the applicability of \ourmethod in reducing redundancy of subspace clustering, as well as our hypothesis. Further study will have to show to what extent \ourmethod can be used further to discover interestingly different sub-parts of a data set.

\section{Conclusion} \label{sec:conc}

We studied the problem of analyzing the relations of two random vectors of potentially different number of attributes, realizations, and possibly not even having a joint distribution. We proposed \ourmethod, a formalization of this problem, and three practical instantiations. We also introduced two alternative formulations that provide a wider gamut to tackle the problem. Experimental results on various tasks evidence the potential of our methods. For future work, we plan to look at non-linear transformations using kernel methods and deep learning. We are investigating how to measure privacy preservation and data security between original and released data using \ourmethod.

\section*{Acknowledgements}
The authors are supported by the Cluster of Excellence ``Multimodal Computing and Interaction'' within the Excellence Initiative of the German Federal Government.

\bibliographystyle{abbrv}
\bibliography{citation}

\begin{thebibliography}{10}
\bibitem{pearson}
K.~Pearson.
\newblock On the criterion that a given system of deviations from the probable
  in the case of a correlated system of variables is such that it can be
  reasonably supposed to have arisen from random sampling.
\newblock {\em Philosophical Magazine}, 50:157--175, 1900.
\end{thebibliography}

\newpage

\newif\ifapx
\apxtrue

\ifapx
\appendix

\section{Proofs} \label{sec:proofs}

\begin{proof} (\textbf{Lemma~\ref{lem:mallows}})
We prove for the case when $\SizeX \neq \SizeY$. The proof for when $\SizeX = \SizeY$ follows similarly. W.l.o.g., we assume that $\SizeX < \SizeY$.

As aforementioned, since $\SizeX \neq \SizeY$, we replicate each realization such that both distributions have $\SizeX \times \SizeY$ realizations. We write the new sets of realizations as $\{\x'_1, \ldots, \x'_{\SizeX \times \SizeY}\}$ and $\{\y'_1, \ldots, \y'_{\SizeX \times \SizeY}\}$. We have

\begin{equation} \label{eq:samplemd}
\diff(p(\X), q(\Y)) \sim \left(\sum_{\{j_1, \ldots, j_{\SizeX \times \SizeY}\}} \sum_{i=1}^{\SizeX \times \SizeY} |\x'_i - \y'_{j_i}|^t\right)^{1/t}
\end{equation}
where $\{j_1, \ldots, j_{\SizeX \times \SizeY}\}$ is a permutation of $\{1, \ldots, \SizeX \times \SizeY\}$. For each $i \in [1, \SizeX]$, and each $j \in [1, \SizeY]$, the term $|\x_i - \y_j|^t$ appears at most $\min(\SizeX, \SizeY) = \SizeX$ in a single permutation. Thus, the total number of times $|\x_i - \y_j|^t$ appears in the sum is
$$\sum_{c=1}^\SizeX c (\SizeX \times \SizeY - \SizeX)! \prod\limits_{r=0}^{\SizeX-c-1} (\SizeX \times \SizeY - \SizeY - r).$$
As we make no preference in picking $i$ and $j$, every term $|\x_i - \y_j|^k$ appears the same number of times as specified above. Hence, we arrive at the result.
\end{proof}

\begin{proof} (\textbf{Lemma~\ref{lem:beta}})
From $\vecu^T \vecu = \vecv^T \vecv = 1$, it holds that $|\vecu^T \Xb| \leq \sqrt{\bar{\DimX}}$ and $|\vecv^T \Yb| \leq \sqrt{\bar{\DimY}}$. Therefore, with $\beta = \sqrt{\frac{\bar{\DimX}}{\bar{\DimY}}}$, we have that $\beta \vecv^T \Yb \leq \sqrt{\bar{\DimX}}$.
\end{proof}

\begin{proof} (\textbf{Lemma~\ref{lem:beta2}})
Let $\vecu' \in \mathbb{R}^{\DimX + c}$ be such that $\vecu' = [\vecu, \mathbf{0}]$ and $\vecv' \in \mathbb{R}^{\DimY + d}$ be such that $\vecv' = [\vecv, \mathbf{0}]$. We have that $(\vecu')^T \vecu' = (\vecv')^T \vecv' = 1$ and
$$\diff\left(p((\vecu')^T \Xb), q(\beta (\vecv')^T \Yb)\right) = \diff(p(\vecu^T \Xb), q(\beta\vecv^T \Yb)).$$
That is, $\diff\left(p((\vecu')^T \Xb), q(\beta (\vecv')^T \Yb)\right)$ is small, i.e.\ $(\vecu', \vecv')$ and hence $(\vecu, \vecv)$ are identifiable by \ourmethod.
\end{proof}

\begin{proof} (\textbf{Proposition~\ref{prop:recda}})
Our proof is based on~\cite{le:ica}. In particular, we have
\begin{align*}
& \frac{\lambda}{\SizeX} \sum_{i=1}^\SizeX ||\vecu \vecu^T \tilde{\xb_i} - \tilde{\xb_i}||^2\\
& = \lambda \times \mathit{tr}\left[(\vecu \vecu^T - \mathbf{I})^T (\vecu \vecu^T - \mathbf{I}) \times \frac{1}{\SizeX} \sum_{i=1}^\SizeX \tilde{\xb_i} (\tilde{\xb_i})^T \right]
\end{align*}
As $\Xb$ is whitened, $\frac{1}{\SizeX} \sum_{i=1}^\SizeX \tilde{\xb_i} (\tilde{\xb_i})^T = \mathbf{I}$. Hence,
$$\frac{\lambda}{\SizeX} \sum_{i=1}^\SizeX ||\vecu \vecu^T \tilde{\xb_i} - \tilde{\xb_i}||^2
= ||\vecu \vecu^T - \mathbf{I}||_F^2.$$
\end{proof}

\begin{proof} (\textbf{Proposition~\ref{prop:recda2}})
Our proof again is based on~\cite{le:ica}. In particular, we have
\begin{align*}
& (\vecu^T \vecu - 1)^2\\
& = (\vecu^T \vecu - 1)(\vecu^T \vecu - 1)\\
& = \vecu^T \vecu \vecu^T \vecu - 2 \vecu^T \vecu + 1\\
& = \mathit{tr}(\vecu \vecu^T \vecu \vecu^T) - 2\mathit{tr}(\vecu \vecu^T) + \mathit{tr}(\mathbf{I}_\DimX) + 1 - \DimX\\
& = \mathit{tr}\left[\vecu \vecu^T \vecu \vecu^T - 2 \vecu \vecu^T + \mathbf{I}_\DimX\right] + 1 - \DimX\\
& = \mathit{tr}\left[(\vecu \vecu^T - \mathbf{I}_\DimX)^T (\vecu \vecu^T - \mathbf{I}_\DimX)\right] + 1 - \DimX\\
& = ||\vecu \vecu^T - \mathbf{I}_\DimX||_F^2 + 1 - \DimX.
\end{align*}
\end{proof}

\begin{proof} (\textbf{Lemma~\ref{lem:recda}})
From Propositions~\ref{prop:recda} and~\ref{prop:recda2}, we can see that when $\Xb$ and $\Yb$ are whitened, \Rcda is equivalent to
\begin{align*}
\arg\min_{\substack{\vecu^T \vecu^j = 0, \forall j \in [1, i-1]\\\vecv^T \vecv^j = 0, \forall j \in [1, i-1]}} & \left[\lambda (\vecu^T \vecu - 1)^2 + \delta (\vecv^T \vecv - 1)^2 \right.\\
& \left. + \diff\left(p(\vecu^T \Xb), q(\beta\vecv^T \Yb)\right)\right].
\end{align*}
When $\lambda \uparrow +\infty$ and $\delta \uparrow +\infty$, the unit norm costs reduce to hard constraints.
\end{proof}

\section{Sample Estimator of the Quadratic Measure} \label{sec:quadest}

We have $\diff(p(Z), q(Z))$ is equal to
\begin{align*}
\diff(p(Z), q(Z)) = & \int p(z) p(z) dz + \int q(z) q(z) dz \\
& - \int q(z) p(z) dz - \int p(z) q(z) dz
\end{align*}
which is $E_p(p(Z)) + E_q(q(Z)) - E_p(q(Z)) - E_q(p(Z))$.

With \kde, $\diff(p(Z), q(Z))$ becomes
\begin{align*}
\frac{1}{\SizeX} \sum_{i=1}^\SizeX \widehat{p}_{h}(x_i) + \frac{1}{\SizeY} \sum_{j=1}^\SizeY \widehat{q}_{b}(y_j) \\
- \frac{1}{\SizeX} \sum_{i=1}^\SizeX \widehat{q}_{b}(\x_i) - \frac{1}{\SizeY} \sum_{j=1}^\SizeY \widehat{p}_{h}(y_j)
\end{align*}
Thus, we arrive at the result.

\section{Multi-dimensional Reconstruction \ourmethod} \label{sec:mrcda}

We form this alternative formulation by combining \Mcda and \Rcda. Formally, it is defined as follow.

\begin{definition} \label{def:mrcda}
Multi-dimensional \Rcda (\MRcda) solves for $\U = (\vecu^1, \ldots, \vecu^r) \in \mathbb{R}^{\DimX \times r}$ and $\V = (\vecv^1, \ldots, \vecv^r) \in \mathbb{R}^{\DimY \times r}$ where $(\U, \V)$ is the solution of
\begin{align*}
& \arg\min_{\U^T \U = \V^T \V = \mathbf{I}_r} \left(\frac{\lambda}{\SizeX} \sum_{i=1}^\SizeX ||\U \U^T \tilde{\xb_i} - \tilde{\xb_i}||^2 \right.\\
& \left. + \frac{\delta}{\SizeY} \sum_{j=1}^\SizeY ||\V \V^T \tilde{\yb_j} - \tilde{\yb_j}||^2 + \diff\left(p(\U^T \Xb), q(\mathbf{\Gamma} \V^T \Yb)\right)\right)
\end{align*}
with $\tilde{\xb_i} = \xb_i - \mu(\Xb_i)$ being the centered version of $\xb_i$, $\tilde{\yb_j} = \yb_j - \mu(\Yb_j)$ being the centered version of $\yb_j$, $p(.)$ and $q(.)$ being pdfs, $\diff$ being a divergence measure of pdfs, $\lambda > 0$ and $\delta > 0$ being weights of reconstruction costs, and $\mathbf{\Gamma} = \mathit{diag}(\beta_1, \ldots, \beta_r)$ being the scaling matrix to bring $(\vecu^i)^T \Xb$ and $(\vecv^i)^T \Yb$ to the same domain.
\end{definition}

To solve \MRcda, we again use \pe divergence. Note that for both \Mcda and \MRcda, the bandwidths are: $\sigma_\Xb = r \times \mathit{median}\{||\xb_i - \xb_j||: 1 \leq i, j \leq \SizeX\}$ and $\sigma_\Yb = \left(\sum_{k=1}^r \beta_i\right) \times \mathit{median}\{||(\yb_i - \yb_j)||: 1 \leq i, j \leq \SizeY\}$.

\section{Sensitivity to Noisy Attributes} \label{sec:noise}

To assess quality of all methods tested to noisy attributes, we extend our setup in Section~\ref{sec:retrieve}. In particular, we first generate $\SizeX$ samples for $\Xb \in \mathbb{R}^{5 + c}$ where $c \geq 2$ and $\Xb_i \sim \mathcal{N}(0, 1)$. Then we generate $\SizeY$ samples for $\Yb \in \mathbb{R}^{3 + c}$ where $\Yb_1$ to $\Yb_4$ follow the three types of relations: linear, mixed, and non-linear; in addition, $\Yb_j \sim \mathcal{N}(0, 1)$ for $j \in [5, 3 + c]$. Hence, $c$ controls the number of noisy attributes. For exposition, in Table~\ref{tab:noise} we present results on the setting: $\SizeX = \SizeY$, row order intact, and non-linear relations. We can see that our methods are relatively stable to noisy dimensions, which is in line with our theoretical analysis in Section~\ref{sec:beta}. This highlights the benefits of our setting of $\beta$.

\begin{table*}[tb]
\centering
\small
\begin{tabular}{lrrrrrrrr}
\toprule
& \multicolumn{3}{l}{\textbf{\cda}} & & & & &\\
\cmidrule(r){2-4}
$c$ & \textbf{\qcda} & \textbf{\mRcda} & \textbf{\peMRcda} & \textbf{\cca} & \textbf{\lscca} & \textbf{\scl} & \textbf{\dac} & \textbf{\ssc}\\

\otoprule

2 & \textbf{0.36} & \textbf{0.35} & \textbf{0.35} & 0.50 & 0.48 & 0.52 & 0.57 & 0.54\\

4 & 0.38 & \textbf{0.36} & \textbf{0.36} & 0.53 & 0.51 & 0.55 & 0.59 & 0.56\\

6 & \textbf{0.35} & 0.38 & \textbf{0.36} & 0.53 & 0.53 & 0.53 & 0.63 & 0.59\\

8 & 0.38 & \textbf{0.36} & \textbf{0.37} & 0.56 & 0.56 & 0.56 & 0.66 & 0.60\\

10 & \textbf{0.39} & \textbf{0.38} & \textbf{0.39} & 0.58 & 0.56 & 0.58 & 0.67 & 0.61\\

\bottomrule
\end{tabular}
\caption{[Lower is better] Frobenius norm errors against the number of noisy attributes for the setting: $\SizeX = \SizeY$, row order intact, and non-linear relations. All values are obtained after 10 trials. Best values and comparable ones according to the Wilcoxon signed rank test at $\alpha = 5\%$ are in \textbf{bold}.} \label{tab:noise} 
\end{table*}

\section{Optimizing vs.\ Not Optimizing $\beta$} \label{sec:optimizing}

We use the setup in Section~\ref{sec:retrieve} to compare \ourmethod methods with and without optimizing $\beta$. The results are in Table~\ref{tab:compare}. Overall, we can see that there is no clear winner between the two options. This could be explained as follows. When $\beta$ is a variable to be optimized, the search space and hence the optimization problem become more complex, especially for \qcda and \pecda. With a complex search space the optimizers employed may be more prone to local optimum. This explains why optimizing $\beta$ currently does not bring a clear improvement, while costing more time. In particular, optimizing $\beta$ incurs from 8\% to 15\% more time than iteratively setting it; please see the last line of Table~\ref{tab:compare} for more details.

\begin{table*}[tb]
\centering
\small
\begin{tabular}{llrrrrrr}
\toprule
& & \multicolumn{3}{l}{\textbf{Iteratively Setting $\beta$}} & \multicolumn{3}{l}{\textbf{Optimizing $\beta$}}\\
\cmidrule(r){3-5} \cmidrule(lr){6-8}

Setting & \textbf{Type} & \textbf{\qcda} & \textbf{\mRcda} & \textbf{\peMRcda} & \textbf{*\qcda} & \textbf{*\mRcda} & \textbf{*\peMRcda}\\

\otoprule

\multirow{2}{*}{\emph{$\SizeX = \SizeY$}} & Linear & 0.31 & \textbf{0.30} & 0.31 & 0.33 & \textbf{0.29} & 0.32\\

& Mixed & \textbf{0.34} & \textbf{0.34} & \textbf{0.35} & \textbf{0.35} & 0.36 & \textbf{0.34}\\

& Non-linear & 0.36 & \textbf{0.35} & \textbf{0.35} & \textbf{0.34} & \textbf{0.35} & \textbf{0.34}\\

\midrule
\midrule

\multirow{3}{*}{\emph{$\SizeX \neq \SizeY$}} & Linear & 0.33 & \textbf{0.31} & \textbf{0.30} & 0.33 & 0.34 & \textbf{0.31}\\

& Mixed & \textbf{0.36} & \textbf{0.36} & 0.38 & \textbf{0.35} & 0.37 & 0.37\\

& Non-linear & \textbf{0.39} & \textbf{0.38} & \textbf{0.39} & 0.40 & 0.41 & 0.40\\

\midrule

\textbf{Average runtime} &  & 1473 & 217 & 284 & 1611 & 252 & 319\\

\bottomrule
\end{tabular}
\caption{[Lower is better] Frobenius norm errors for iteratively setting $\beta$ vs.\ optimizing it. All values are obtained after 10 trials. Best values and comparable ones according to the Wilcoxon signed rank test at $\alpha = 5\%$ are in \textbf{bold}. The superscript `*' indicates \ourmethod methods that optimize $\beta$. The last line is the average runtime (in seconds) of all methods.} \label{tab:compare} 
\end{table*}

\section{Efficiency} \label{sec:time}

Above we show that iteratively setting $\beta$ gives better efficiency than optimizing it. There is another factor that affects the runtime, which is the reconstruction costs. To study this, we use the runtime obtained from our experiments in Section~\ref{sec:retrieve}. We summarize the results in Table~\ref{tab:recontime}. Overall, we observe that using reconstruction costs bring about 5 times speedup compared to the other variants. This could be attributed to the fact that \ourmethod with reconstruction costs yields unconstrained optimization problems where fast unconstrained optimizer, like L-BFGS, can be applied.

\begin{table*}[tb]
\centering
\small
\begin{tabular}{rrrrrr}
\toprule
\multicolumn{3}{l}{\textbf{No reconstruction costs}} & \multicolumn{3}{l}{\textbf{Reconstruction costs}}\\
\cmidrule(r){1-3} \cmidrule(lr){4-6}

\textbf{\qcda} & \textbf{\mcda} & \textbf{\peMcda} & \textbf{\qRcda} & \textbf{\mRcda} & \textbf{\peMRcda}\\

\otoprule
		
1473 & 1265 & 1622 & 245 & 217 & 284\\

\bottomrule
\end{tabular}
\caption{[Lower is better] Average runtime (in seconds) of \ourmethod methods not using reconstruction costs and the respective ones using reconstruction costs.} \label{tab:recontime} 
\end{table*}

\if

\end{document}